\documentclass[fleqn,10pt,twocolumn]{AROB}
\usepackage{graphicx}
\usepackage{ascmac}
\usepackage{amsmath}
\usepackage{float} 
\usepackage{afterpage}

\title{LLM-mediated Dynamic Plan Generation with a Multi-Agent Approach}

\author{Reo Abe${}^{1\dagger}$, Akifumi Ito${}^{2}$, Kanata Takayasu${}^{2}$ and Satoshi Kurihara${}^{3}$}
\speaker{Reo Abe}

\affils{${}^{1}$University of Keio, Kanagawa, Japan\\
(E-mail: reo1431@keio.jp)\\
${}^{2}$University of Keio, Kanagawa, Japan\\
${}^{3}$University of Keio, Kanagawa, Japan\\
(E-mail: satoshi@keio.jp)\\}

\abstract{%
Planning methods with high adaptability to dynamic environments are crucial for the development of autonomous and versatile robots. We propose a method for leveraging a large language model (GPT-4o) to automatically generate networks capable of adapting to dynamic environments. The proposed method collects environmental "status," representing conditions and goals, and uses them to generate agents. These agents are interconnected on the basis of specific conditions, resulting in networks that combine flexibility and generality. We conducted evaluation experiments to compare the networks automatically generated with the proposed method with manually constructed ones, confirming the comprehensiveness of the proposed method's networks and their higher generality. This research marks a significant advancement toward the development of versatile planning methods applicable to robotics, autonomous vehicles, smart systems, and other complex environments.  }


\begin{document}

\maketitle

%-----------------------------------------------------------------------

\section{Introduction}
Autonomous robot planning for coexistence in daily life with high adaptability to dynamic environments is crucial for the development of autonomous and versatile robots. In such environments, a balance between reactive planning, which enables quick action selection, and deliberative planning, which requires time to determine a sequence of actions to achieve a goal, is essential. Reactive planning is effective in rapidly changing situations in which quick decisions are needed, while deliberative planning enables careful responses in unforeseen circumstances for achieving complex objectives.

Planning methods based on machine learning (ML) and large language models (LLMs) have been gaining attention \cite{levine2016visuomotor,karnchanachari2024nuplan,chatzilygeroudis2020model,zeng2020transporter,HiTS2021}. ML-based planning excels in quickly selecting actions on the basis of the environment, making it suitable for scenarios that require immediate responses. However, these methods rely heavily on pre-trained models, making it challenging to adapt flexibly to unexpected situations.

In contrast, LLM-based planning leverages the extensive knowledge contained within LLMs to generate diverse goals and action sequences on the basis of environmental information. This flexibility enables LLMs to adapt to both new and unstructured environments. However, many studies that have applied LLMs to robot planning assumed integration with existing manually constructed planning frameworks, which limits the full utilization of LLMs' broad representational capacity \cite{ahn2022saycan,huang2023scene,rt2,zeng2023survey,embodied_task_planning}. Therefore, a method for maximizing the potential of LLMs for more effective application in robot planning is required.

To address this issue, we propose a method that leverages LLMs to derive specific plan sequences from natural language instructions without relying on existing robot planning frameworks. By making the planner through a multi-agent approach, the proposed method incorporates both reactivity and deliberation. With this multi-agent planning approach, each operand in a STRIPS ()-like structure that functions as an autonomous agent, and through an activation spreading mechanism, these agents coordinate to use quick responses and careful decision-making depending on the situation  \cite{strips}.

With the proposed method, an LLM is used to generate various sentences about human actions on the basis of conditions such as "person," "location," and "time of day," and the purpose of each action is extracted. For each action, the necessary sub-actions are then recursively generated to achieve that purpose, automatically obtaining an action sequence to attain the goal. Conventional studies relied on existing robot-planning frameworks due to the manual nature of planning construction, but the proposed method automates this process, enabling the derivation and execution of diverse plans.

The proposed method facilitates the development of flexible and scalable planning systems in robotics and has the potential to be applied in other fields requiring dynamic, real-time decision-making such as smart systems and autonomous avatars in the metaverse.

%%%%%%%%%%Related Work%%%%%%%%%%
\section{Related Work}

\subsection{Robot Planning Using Machine Learning}  
ML-based robot planning provides the ability for robots to quickly select actions in response to their environments, making it suitable for scenarios requiring reactive responses. Levine et al. (2016) proposed a method that uses deep reinforcement learning for robotic control, achieving high performance in specific tasks \cite{levine2016visuomotor}. However, such methods heavily rely on pre-trained models, making it challenging to adapt to unexpected situations and unknown environments.

Further studies, such as on transporter networks \cite{zeng2020transporter} and model-based reinforcement learning \cite{chatzilygeroudis2020model}, were conducted to enhance robotic manipulation efficiency. Despite their contributions, these studies faced limitations in terms of their methods fully adapting to dynamic and unpredictable environments.

To address these challenges, the nuPlan benchmark \cite{karnchanachari2024nuplan} provides a comprehensive framework for learning-based planning. It particularly targets the adaptation capabilities of autonomous vehicles in real-world scenarios, offering a new evaluation framework to assess planning performance in dynamic environments. While nuPlan demonstrates adaptability to dynamic environments, its planning accuracy decreases in unpredictable situations or previously unseen scenarios.

Similarly, Hierarchical Reinforcement Learning with Timed Subgoals (HiTS) \cite{HiTS2021} introduces an approach for efficient planning in dynamic environments through hierarchical reinforcement learning. By using timed subgoals, HiTS improves the efficiency of the planning processes. However, like nuPlan, HiTS faces challenges in maintaining planning accuracy in unpredictable environments or scenarios not anticipated during training.

\subsection{Robot Planning Using LLMs}
Advancements in robot planning have highlighted the potential of LLMs. LLMs, with their vast knowledge base, can generate diverse sequences of goals and actions on the basis of environmental inputs, making them highly adaptable to novel and unstructured environments \cite{ahn2022saycan}\cite{huang2023scene}. For instance, Google Research's SayCan \cite{ahn2022saycan} demonstrated the ability of robots to execute tasks on the basis of natural language instructions, showcasing the integration potential of LLMs and robotics.

RT-2 (Robotic Transformer 2) further advances LLM utilization by integrating vision and language data for robot action generation \cite{rt2}. RT-2 leverages large-scale web-based visual and linguistic data to enable flexible responses to unknown environments and objects. 

Zeng et al. (2023) conducted a comprehensive survey, "Large Language Models for Robotics: A Survey" \cite{zeng2023survey}, analyzing the use of LLMs in robotics. This survey identified both the potential and limitations of LLM-based methods, noting that most studies focus on integrating LLMs with existing planning frameworks. However, methods for fully using the generalization and adaptability of LLMs independently require further research.

Task Planning Agent (TaPA) introduces a novel approach by integrating LLMs with visual-recognition models to generate action plans on the basis of objects in the environment \cite{embodied_task_planning}. While TaPA has demonstrated effectiveness in executing complex tasks and real-world robot operations, it shares common challenges with SayCan and RT-2 in leveraging LLMs' representational capabilities to achieve standalone generalization and adaptability. Addressing these limitations remains a critical research goal.

%%%%%%%%%%Proposed Method%%%%%%%%%%
\section{Proposed Method}
\begin{figure}[H]
\begin{center}
\includegraphics[width=8cm]{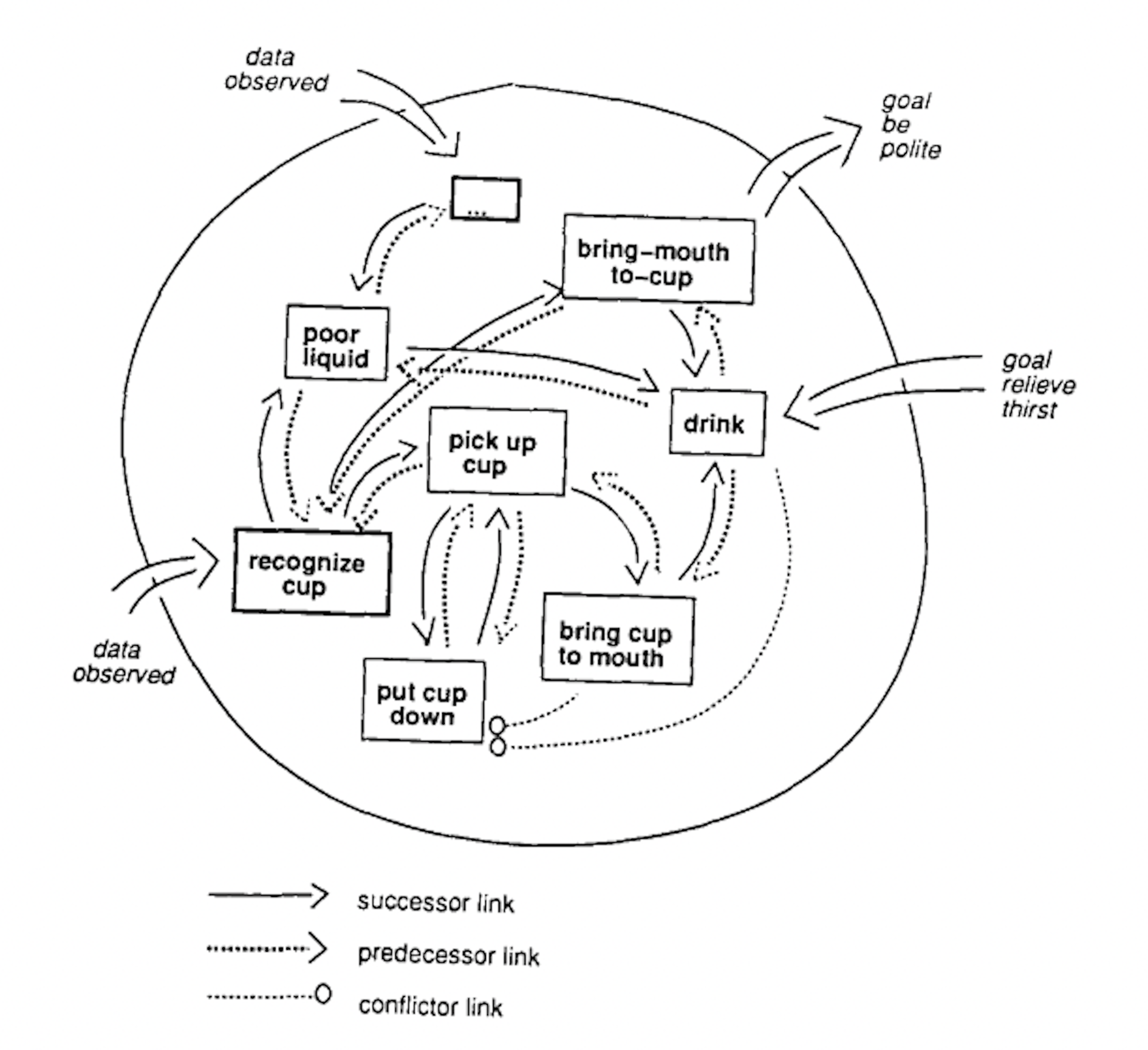}
\caption{\label{fig:ana} ANA Network}
\end{center}
\end{figure}

The proposed method automatically generates networks adaptable to dynamic environments by leveraging the Agent Network Architecture (ANA) \cite{ANA1}\cite{ANA2} and a large language model (GPT-4o, hereafter referred to as GPT). ANA is an architecture designed to enable planning by integrating both reactive and deliberative planning capabilities in dynamic and unpredictable environments. It uses a network of multiple autonomous agents \cite{agent} that collaborate to perform tasks. An example of an ANA network is shown in Figure \ref{fig:ana}.

In ANA network, each agent operates as an independent "operand," maintaining specific relationships with the environment while selecting actions on the basis of the current situation. The ANA's central mechanism, activity propagation, enables real-time information exchange between agents, enabling flexible and scalable planning in dynamic environments. Through activity propagation, the ANA achieves a balance between reactive planning (rapid responses to sudden changes) and deliberative planning (constructing action sequences to achieve long-term goals).

However, ANA faces the following challenges:

\begin{itemize}
    \item \textbf{Cost of Agent Design:}  
    Each agent functions as a specific "operand," requiring the definition of a "Condition list," "Add list," and "Delete list." These design tasks are often conducted manually, which becomes increasingly costly as the number of agents grows in large-scale networks. This leads to an exponential increase in design costs.

    \item \textbf{Difficulty in Scaling Up:}  
    Scaling up the network requires proportional increases in agent design efforts, making it highly challenging to extend the system to larger scales. Optimizing interactions between agents in large-scale networks also remains an unresolved issue, which may result in degraded performance in complex environments.
\end{itemize}

\begin{figure}[H]
\begin{center}
\includegraphics[width=8cm]{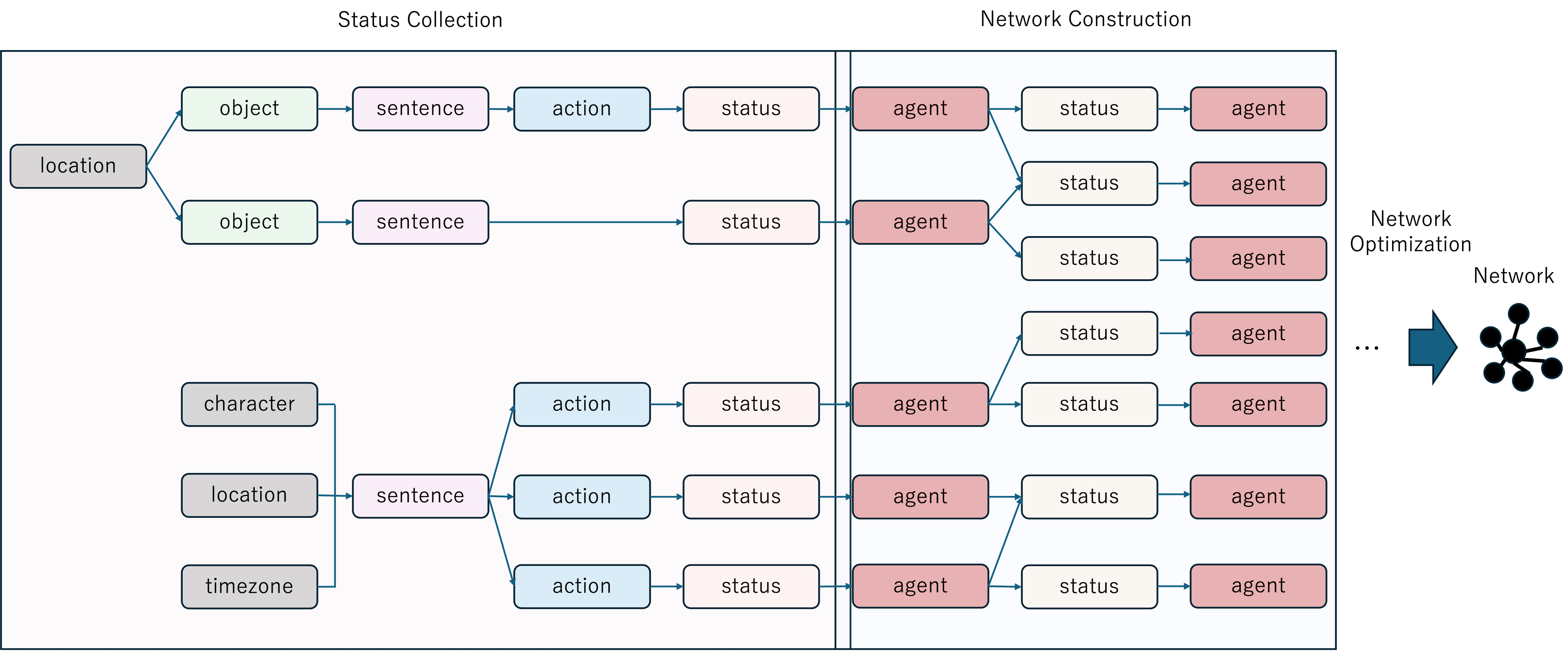}
\caption{\label{fig:method_overview} method overview}
\end{center}
\end{figure}

To address these challenges, the proposed method automates the generation of agents using GPT while retaining the ANA as the foundational structure. This reduces design costs while improving the overall flexibility and efficiency of the network. An overview of the proposed method is illustrated in Figure~\ref{fig:method_overview}.

\subsection{Automatic Network Generation}
\begin{figure}[H]
\begin{center}
\includegraphics[width=8cm]{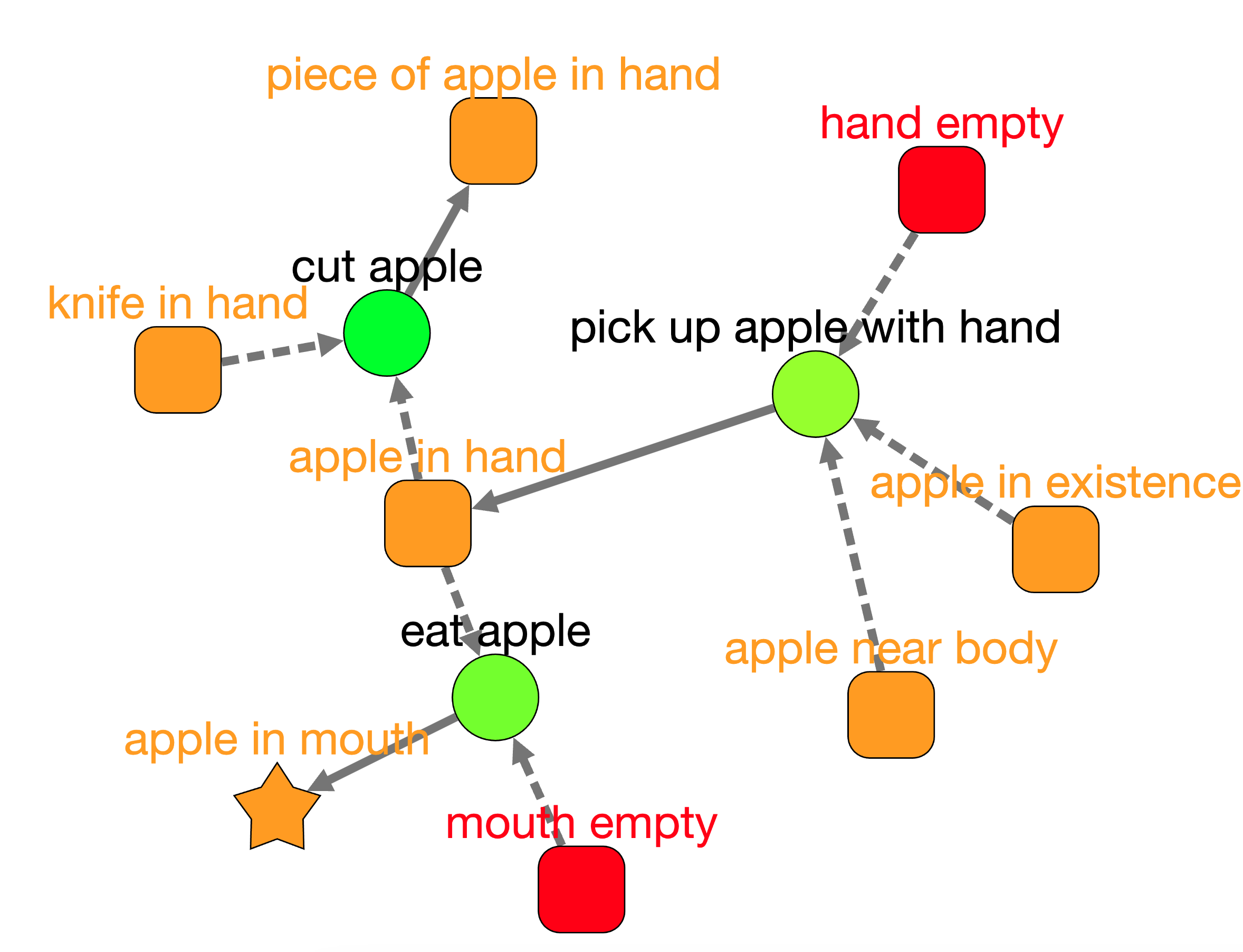}
\caption{\label{fig:sample network} sample network}
\end{center}
\end{figure}

A portion of an automatically generated network is illustrated in Figure\ref{fig:sample network}. The proposed method generates agents required to achieve specific statuses and organizes them into a network. The agents that constitute the network are as follows.

\begin{itembox}[l]{Example of Agent}
Agent: pick up cup \\
Add list: cup in hand \\
Condition list: hand empty, cup near body \\
Delete list: hand empty \\
\\
Agent: pour water \\
Add list: water in cup \\
Condition list: cup empty, cup in hand \\
Delete list: cup empty
\end{itembox}

Each agent consists of three lists: \textit{add list}, \textit{condition list} and \textit{delete list}, which are based on the foundational structure of the Agent Network Architecture (ANA). These lists serve the following purposes:

\begin{itemize}
    \item \textbf{Add list:} A list of statuses added to the environment when the agent is executed.
    \item \textbf{Condition list:} A list of statuses that must be satisfied for the agent to execute.
    \item \textbf{Delete list:} A list of statuses removed from the environment when the agent is executed.
\end{itemize}

Using these lists, interactions between agents are defined, forming a structure that dynamically supports planning across the network. When an agent is executed, it triggers a corresponding primitive action in the environment. For instance, the agent "pick up cup" represents the physical action of a robot picking up a cup, resulting in changes to the environment's state.

\subsection{Overview of Network Generation}
 The method consists of the following three stages:

\begin{enumerate}
    \item \textbf{Status Collection:}
    \begin{itemize}
        \item \textit{Object-Based Method:} Collect statuses on the basis of the state or position of objects.
        \item \textit{Condition-Based Method:} Collect actions on the basis of persons, locations, and time frames.
    \end{itemize}
    \item \textbf{Network Construction:} Generate agents in reverse order starting from the statuses using GPT, and form a connected network.
    \item \textbf{Network Optimization:} Merge similar agents and statuses using GPT to eliminate redundancy and improve efficiency.
\end{enumerate}

Status collection is executed using a combination of the object-based and condition-based methods.

\subsubsection{Object-Based Status Collection}

This method collects statuses on the basis of the state or location of objects. First, GPT is used to list objects present in a specified location or related to it. Next, GPT generates sentences that include these objects as direct objects, which are then organized into the format "verb + object." From these sentences, statuses are generated. GPT also generates spatial information in the format "object + preposition + object", which is directly adopted as statuses.

This method is essential for reflecting the diversity of objects and enhancing network flexibility. It is particularly effective for accommodating new objects, enabling easy adaptation to changes.

\subsubsection{Condition-Based Status Collection}

This method collects human actions on the basis of conditions such as persons, locations, and time frames. GPT is used to generate sentences describing actions under specific conditions, which are then organized into the format "verb + object." From these actions, statuses are generated.

This method complements the object-based method by addressing actions that do not involve direct objects, such as "sleep" or "wait," thus increasing the network's versatility.

\subsection{Network Construction}

The construction of a network begins with the statuses collected in the previous stage. Agents are generated in reverse order starting from these statuses, ensuring the connectivity and effectiveness of the network. A detailed illustration of this process is shown in Figure\ref{fig:generate_network}. The procedure is as follows:

\begin{figure}[H]
\begin{center}
\includegraphics[width=8cm]{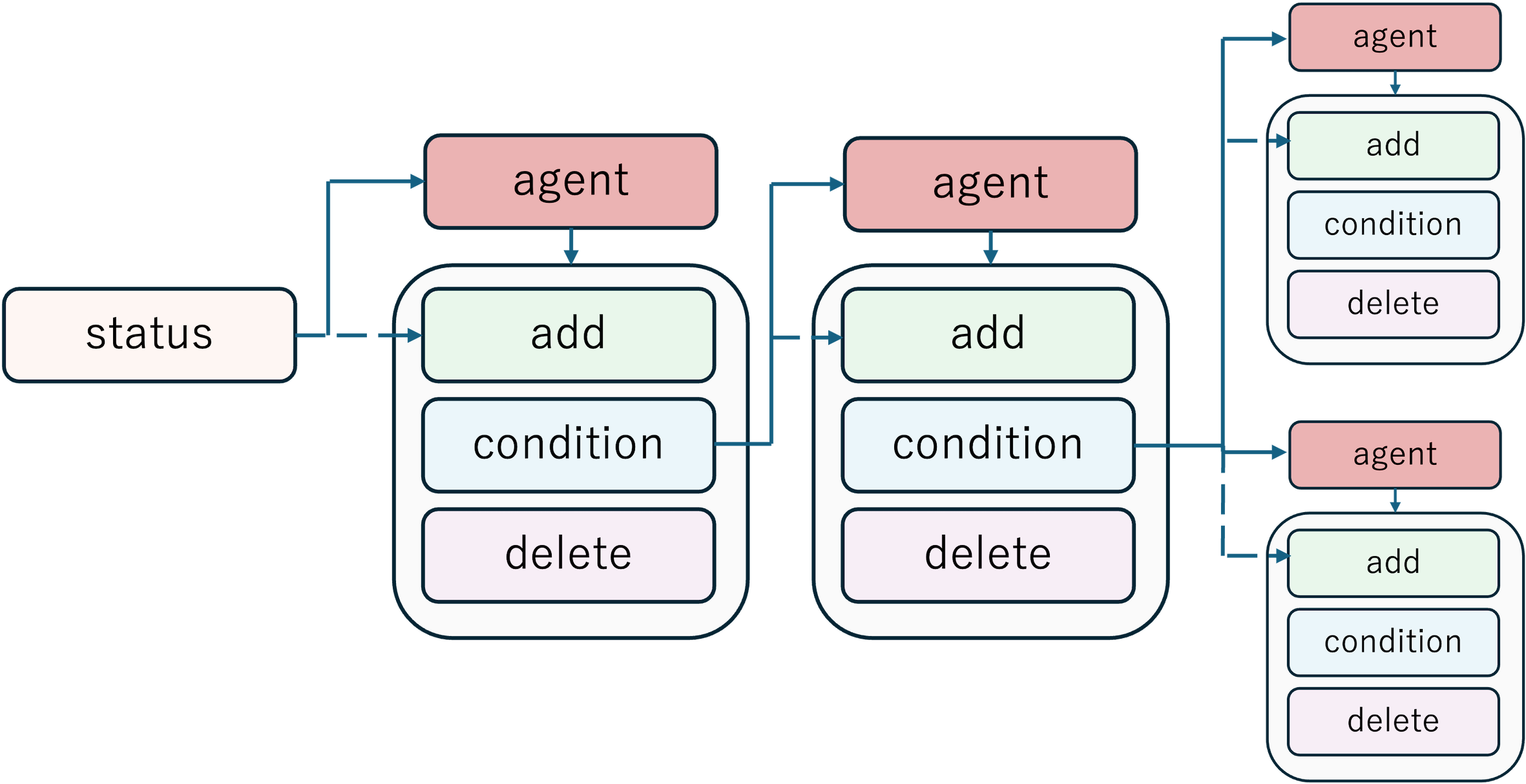}
\caption{\label{fig:generate_network} Generate network}
\end{center}
\end{figure}

\begin{enumerate}
    \item \textbf{Initialization of Starting Statuses:} Network generation begins with the statuses collected during the "Status Collection" phase.
    \item \textbf{Agent Generation:} For each starting status, an agent capable of achieving that status is generated. The agent includes:
    \begin{itemize}
        \item \textbf{Add list:} Includes the starting status and other statuses added to the environment upon execution.
        \item \textbf{Condition list:} Specifies the prerequisites (statuses) required for the agent to execute.
        \item \textbf{Delete list:} Lists statuses removed from the environment upon execution.
    \end{itemize}
    \item \textbf{Processing the Condition List:} For each status in the condition list of the generated agent, new agents are created to fulfill those statuses. This process is repeated until all statuses in the condition list are processed.
    \item \textbf{Recursive Agent Generation:} The same procedure is applied recursively to the condition lists of newly generated agents, expanding the network.
    
    \begin{itembox}[l]{Termination Conditions}
    \begin{itemize}
        \item An agent containing the status in its add list has already been generated.\\
        \item The status is "hand empty": Generate the agent "put down X" and terminate.\\
        \item The status is "X in hand": Generate the agent "pick up X" and terminate.\\
        \item The status is "X near body ": Generate the agent "move to X" and terminate.
    \end{itemize}
    \end{itembox}
\end{enumerate}

\begin{itembox}[l]{Generated Agents}
verb + object: clean window\\
starting status: window clean\\
\\
Agent: clean window with towel \\
Add list: window clean \\
Condition list: towel in hand, window near body \\
Delete list: none \\
\\
Agent: pick up towel \\
Add list: towel in hand \\
Condition list: hand empty, towel near body \\
Delete list: hand empty\\
\\
Agent: move to window \\
Add list: window near body \\
Condition list: window far from body \\
Delete list: window far from body \\
\\
Agent: move to towel \\
Add list: towel near body \\
Condition list: towel far from body \\
Delete list: towel far from body
\end{itembox}
\subsection{Network Optimization}

To eliminate redundancy in the generated network, similar agents and statuses are merged using GPT. First, statuses are evaluated for similarity on the basis of the objects they reference, and GPT assesses their compatibility for merging. Next, agents with identical 'condition list,' 'add list,' and 'delete list' are evaluated for similarity using GPT and merged accordingly.

The optimization process reduces computational load, enhancing efficiency and adaptability in dynamic environments.

\subsection{Significance of the Proposed Method}

The proposed method represents a novel approach that leverages GPT to generate agents and automatically constructs networks on the basis of the foundation of the ANA. This method reduces design costs while enhancing flexibility and scalability in dynamic environments. It also expands the applicability of planning systems to various fields, including robotics, smart systems, and autonomous vehicles.

%%%%%%%%%%Evaluation Method%%%%%%%%%%
\section{Evaluation}

\subsection{Evaluation Methodology}

To validate the effectiveness and efficiency of the proposed method, the following evaluations were conducted:

\subsubsection{Comparison with Manually Constructed Networks}

To evaluate whether the proposed method can generate more comprehensive and generalized networks compared with participants manually constructing networks, we conducted a comparative study consisting of the following steps:

\textbf{Preparation:}  
Eight participants were shown examples of agents and parts of a network automatically generated with the proposed method to help them form a clear image of the networks they were tasked to construct.

\textbf{Network Construction:}  
Participants were provided with a starting status (\textit{window clean}) and instructed to construct networks in two stages:
\begin{itemize}
    \item \textbf{Status Achievement Network:}  
    Construct a minimal network designed solely to achieve the given status.
    \item \textbf{Extended Network:}  
    Extend the status achievement network by increasing the network size while ensuring that the distance from the starting status remains within 6.
\end{itemize}

\textbf{Coverage Calculation:}  
The manually constructed networks were compared with the proposed method’s networks. The coverage was calculated as a quantitative measure to determine the extent to which the proposed method’s network encompasses the manually constructed networks.

\subsubsection{Impact of Network Scale on Planning Success}

To evaluate the impact of the scale of the proposed method’s networks on the success rate of planning, we conducted experiments using networks of different scales. Specifically, planning was executed ten times for each network, and the success rates were measured.

The following three types of networks were used to achieve the goal of "window clean":
\begin{enumerate}
    \item \textbf{Minimal Network:} The smallest network required to achieve "window clean."
    \item \textbf{Distance 5 Network:} A network containing nodes within a distance of 5 from the starting status "window clean."
    \item \textbf{Distance 6 Network:} A network containing nodes within a distance of 6 from the starting status "window clean."
\end{enumerate}

The minimal network was assumed to meet the minimum environmental conditions required to achieve the goal. This network served as the baseline for conducting the planning experiments.

\subsection{Evaluation Results}

Table\ref{tab:evaluation_results} shows the coverage rate of the proposed method’s networks compared with the networks manually constructed by the participants. The networks constructed by the 8 participants contained 29 agents and 49 statuses after removing duplicates.

For calculating the coverage rate, the "Full Scale Network" generated with the proposed method was used. This network contained a total of 3162 agents and 3584 statuses and was primarily centered around specific terms such as "window," "towel," and "water." The coverage rate was calculated using the following formula:

\[
\text{C} = 
\frac{
\text{Number of common agents (statuses)}
}{
\text{Number of agents (statuses) in the participants' network}
}
\]
For this calculation, the representation of agents and statuses in the participants' networks was adjusted to match the representation used in the proposed method's networks. The resulting coverage rates for both agents and statuses were approximately 70\%. This indicates that the proposed method can largely replicate networks constructed by humans. The relatively lower coverage rate was primarily due to the proposed method’s networks being centered around specific terms, which did not fully overlap with the vocabulary used by participants. Expanding the diversity of terms in the proposed method’s networks is expected to improve the coverage rate.

The proposed method’s networks included 529 agents and 752 statuses within a distance of 6 from the starting status "window clean." This demonstrates that the proposed method generates more generalized and comprehensive networks compared with participants manually constructing networks.

\begin{table}[htbp]
\centering
\caption{Coverage rate of proposed method’s networks against participant networks}
\label{tab:evaluation_results}
\begin{tabular}{|l|c|c|}
\hline
\textbf{Network Type}            & \textbf{Agents} & \textbf{Statuses} \\ \hline
Participant Network              & 29              & 49                \\ \hline
Proposed Method’s  Network (Distance 6)    & 529             & 752              \\ \hline
Proposed Method’s  Network (Full Scale)    & 3162            & 3584              \\ \hline
\textbf{Coverage Rate (\%)}      & 72.4\%          & 69.3\%            \\ \hline
\end{tabular}
\end{table}

Table\ref{eval} shows the success rates for achieving the goal under different network scales. The results indicate that while the goal was successfully achieved using the Distance 5 Network, it could not be achieved using the Distance 6 Network. This demonstrates that the scale of the network significantly impacts the success rate of planning.

The results also suggest that a larger network increases the number of possible actions under the same environmental conditions, leading to difficulties in correctly executing the planning process. In large-scale networks, in particular, necessary actions for goal achievement may not be selected, or errors in action selection may occur.

Therefore, it is essential to develop methods for effectively using large-scale networks constructed using LLMs. For instance, selectively expanding only the parts of the network required for goal achievement could help ensure accurate planning. Investigating such approaches remains a critical direction for future work.

\begin{table}[htbp]
\centering
\caption{Planning Success Rates Across Different Network Scales}
\label{eval}
\begin{tabular}{|l|c|c|c|}
\hline
\textbf{Network Type}        & \textbf{Agents} & \textbf{Statuses} & \textbf{Success Rate (\%)} \\ \hline
Minimal Network              & 4               & 7                 & 100                        \\ \hline
Distance 5 Network           & 401             & 619               & 100                        \\ \hline
Distance 6 Network           & 529             & 752               & 0                          \\ \hline
\end{tabular}
\end{table}
\section{Conclusion and Future Work}

We proposed a method for automatically generating networks that are adaptable to dynamic environments using large language models (LLMs). The proposed method uses text collected through LLMs to generate agents and construct networks. An evaluation confirmed that the automatically generated networks minimally encompass the content of networks manually constructed by humans while enabling the construction of larger and more complex networks.

This method demonstrates the potential to reduce the cost of network construction while enabling the automated generation of general-purpose networks that can adapt to dynamic environments. The proposed method also contributes to improving both the efficiency and scalability of planning systems.

However, the evaluation also revealed that as the size of the network increases, planning may become infeasible due to computational constraints. Therefore, it is necessary to develop methods for adjusting the network size during planning, such as selectively expanding only the parts of the network required to achieve specific goals.

Future work will focus on optimizing the utilization of automatically generated networks. Specifically, research will be conducted to develop methods for network scale adjustment and to verify the applicability of the method in real-world dynamic environments. Expanding evaluation methods, such as increasing the number of participants with diverse backgrounds, will also be critical to further validate the generality and effectiveness of the proposed method.

\section{ACKNOWLEDGMENT}
This work was supported by grant NEDO/Technology De-
velopment Project on Next-Generation Artificial Intelligence
Evolving Together With Humans “Constructing Interactive
Story-Type Contents Generation System”

%%%%%%%%%%%%%%%%% BIBLIOGRAPHY IN THE LaTeX file !!!!! %%%%%%%%%%%%%%%%%%%%%%

\end{document}